\documentclass{article}
\usepackage{spconf,amsmath,graphicx}
\usepackage{bm}
\usepackage[subrefformat=parens]{subcaption}
\usepackage{multirow}
\usepackage{amssymb}
\usepackage{cite}
\usepackage{threeparttable}
\usepackage{cite}
\newcommand{\argmax}{\mathop{\rm argmax}\limits}
\newcommand{\argmin}{\mathop{\rm argmin}\limits}
\usepackage{multirow}
\usepackage{amsfonts}

\title{Large-Context Conversational Representation Learning:\\ Self-Supervised Learning for Conversational Documents}

\name{Ryo Masumura, Naoki Makishima, Mana Ihori, Akihiko Takashima, Tomohiro Tanaka, Shota Orihashi}
\address{NTT Media Intelligence Laboratories, NTT Corporation, Japan}
\begin{document}
%
\maketitle

\begin{abstract}
  This paper presents a novel self-supervised learning method for handling conversational documents consisting of transcribed text of human-to-human conversations. One of the key technologies for understanding conversational documents is utterance-level sequential labeling, where labels are estimated from the documents in an utterance-by-utterance manner. The main issue with utterance-level sequential labeling is the difficulty of collecting labeled conversational documents, as manual annotations are very costly. To deal with this issue, we propose large-context conversational representation learning (LC-CRL), a self-supervised learning method specialized for conversational documents. A self-supervised learning task in LC-CRL involves the estimation of an utterance using all the surrounding utterances based on large-context language modeling. In this way, LC-CRL enables us to effectively utilize unlabeled conversational documents and thereby enhances the utterance-level sequential labeling. The results of experiments on scene segmentation tasks using contact center conversational datasets demonstrate the effectiveness of the proposed method.
  \end{abstract}

\section{Introduction}
The development of automatic speech recognition technologies has made it increasingly important to utilize the rich information in human-to-human conversations. One of the most famous applications is the data-mining system used in contact centers, where telephone conversations between an operator and customer are utilized for identifying the customer's needs or a problem with the product. Our objective is to develop natural language processing technologies to linguistically understand conversational documents consisting of transcribed text of these human-to-human conversations.

One of the key technologies for understanding conversational documents is utterance-level sequential labeling, where labels are estimated from conversational documents in an utterance-by-utterance manner. Utterance-level sequential labeling is often utilized for building topic segmentation \cite{tsunoo2017,tsunoo2017_2}, scene segmentation \cite{masumura_apsipa2018,orihashi_interspeech2020}, dialogue act classification \cite{tran2017,kumar2018,chen_sigir2018,jiao_naacl2019,raheja_naacl2019,yu_cikm2019}, and end-of-turn detection \cite{masumura_interspeech2017,masumura_sigdial2018}. To build stable utterance-level sequential labeling methods for the conversational documents, it is essential to precisely capture conversational contexts, i.e., who spoke what in what order. In fact, linguistic information about not only the text but also the speaker labels is often taken into consideration \cite{masumura_apsipa2018,masumura_sigdial2018}. Needless to say, a large amount of labeled conversational documents are required for stable modeling. 

However, it is usually difficult to collect enough labeled conversational documents because the manual annotations are very costly. To address this problem, we focus on self-supervised learning using unlabeled datasets \cite{mikolov_nips2013,pennington_emnlp2014,kiros_nips2015,peters_acl2017,peters_naacl2018,devlin_arxiv2018}. Self-supervised learning is a form of unsupervised learning in which unlabeled data is only utilized for designing supervised training settings. It is usually applied during the pre-training stage of several supervised tasks. We suggest that self-supervised learning is a suitable approach for handling conversational documents because unlabeled conversational documents can be collected by utilizing a large number of speech recognition datasets. The key challenge is how to set the self-supervised task for the conversational documents. 

In this paper, we propose a self-supervised learning for conversational documents called large-context conversational representation learning (LC-CRL). Our concept is to estimate an utterance by using all the surrounding utterances. To this end, we introduce a novel large-context language model, which is an extended model of the forward-backward hierarchical recurrent encoder-decoder \cite{masumura_asru2019}, so that we can estimate not only linguistic information but also speaker information. After performing the self-supervised learning, we utilize the pre-trained network for building state-of-the-art utterance-level sequential labeling based on hierarchical bidirectional long short-term memory recurrent neural network conditional random fields (H-BLSTM-CRF) \cite{kumar2018,chen_sigir2018}. To the best of our knowledge, this is the first study in which self-supervised learning has been applied to conversational documents. In experiments using contact center dialogue datasets, we examine call scene segmentation using utterance-level sequential labeling and demonstrate that out proposed self-supervised learning method yields a better performance than conventional ones. 

Our contributions are summarized as follows.
\begin{itemize}
\item We provide a detailed definition of LC-CRL and its implementation method based on large-context language modeling. We also provide a two-stage representation learning method in which context-dependent word representations, i.e., ELMo \cite{peters_naacl2018}, are trained in the first stage and context-dependent utterance representations, i.e., LC-CRL, are trained in the second stage.
\item We provide a building method of H-BLSTM-CRF-based utterance-level sequential labeling after performing LC-CRL-based pre-training. Our building method enables us to utilize pre-trained networks other than a CRF-based classifier for training the H-BLSTM-CRF. 
\end{itemize}

\section{Related Work}

{\bf Self-supervised learning: } Self-supervised learning is a form of unsupervised learning in which unlabeled data is only utilized for designing the supervised training settings. It is usually utilized in the pre-training stage of several supervised tasks. In natural language processing, word representations and sentence representations are often acquired by self-supervised learning. Initial studies examined context-independent word representations such as continuous bag-of-words, skip-gram \cite{mikolov_nips2013}, and GloVe \cite{pennington_emnlp2014}, while more recent studies have developed context-dependent word representations such as ELMo and BERT \cite{peters_acl2017,peters_naacl2018,devlin_arxiv2018}. Context-independent sentence representations such as skip-thought vector \cite{kiros_nips2015} have also been developed. In this paper, we propose a self-supervised learning method for acquiring context-dependent utterance representations that can consider all past and all future conversational contexts beyond the utterance boundaries. We also present a method that combines the context-dependent utterance representations with context-dependent word representations.

\smallskip
\noindent {\bf Large-context language modeling: } Large-context language models that can efficiently capture long-range linguistic contexts from conversational documents beyond the utterance boundaries have received significant attention in recent studies \cite{lin_emnlp2015,wang_acl2016,liu_icassp2017,masumura_interspeech2018,masumura_interspeech2019,kim_slt2018,pundak_slt2018,masumura_icassp2019,jean_arxiv2017,wang_emnlp2018,maruf_acl2018,serban_aaai2016,serban_aaai2017}. These models can assign generative probabilities to words while considering contexts beyond the utterance boundaries. Several studies have reported that leveraging the context information of past utterances can improve the perplexity and the word error rate \cite{lin_emnlp2015,wang_acl2016,liu_icassp2017,masumura_interspeech2018}. Other studies have shown that large-context end-to-end methods offer a superior performance to utterance-level or sentence-level end-to-end methods in automatic speech recognition \cite{kim_slt2018,pundak_slt2018,masumura_icassp2019}, machine translation \cite{jean_arxiv2017,wang_emnlp2018,maruf_acl2018}, and response generation for dialogue systems \cite{serban_aaai2016,serban_aaai2017}. Furthermore, large-context language models that can consider not only past but also future contexts have been presented \cite{masumura_asru2019}. In this paper, we utilize large-context language models for self-supervised learning specialized to conversational documents. 

\smallskip
\noindent {\bf Utterance-level sequential labeling: } Utterance-level sequential labeling, which estimates utterance-level labels from conversational documents, is used for topic segmentation \cite{tsunoo2017,tsunoo2017_2}, scene segmentation \cite{masumura_apsipa2018}, dialogue act classification \cite{tran2017,kumar2018,chen_sigir2018,jiao_naacl2019,raheja_naacl2019,yu_cikm2019}, and end-of-turn detection \cite{masumura_interspeech2017,masumura_sigdial2018}. Most of these techniques use hierarchical recurrent neural networks in which word-level networks and utterance-level networks are hierarchically structured to capture not only contexts within an utterance but also contexts beyond the utterance boundaries. In this paper, we focus on speaker-aware hierarchical recurrent neural networks that can consider not only textual information but also speaker information \cite{masumura_apsipa2018}. In addition, we combine the speaker-aware hierarchical recurrent neural networks with a CRF layer that can take consistencies between labels into consideration. While the previous studies used only labeled datasets to train the utterance-level sequential labeling, we leverage unlabeled conversational documents for mitigating the data scarcity problem.

\setcounter{table}{0}
\begin{table*}[t!]
 \caption{Example of a conversational document in a contact center dialogue.}
 \small
 \begin{center}
   \begin{tabular}{|lll|} \hline
     & Speaker & Text \\ \hline \hline
     $U_{1}$ & Operator & Thank you for calling XXX Bank, my name is Jerry, how can I help you today? \\
     $U_2$ & Customer & Hi Jerry, \{um\} I lost my wallet \{um\}. \\
     $U_3$ & Customer & And I need to deactivate the card that \{I\} I lost inside there. \\
     $U_4$ &Operator & Oh, that's terrible news, but yeah, I can definitely help you out with that. \\
     $U_5$ &Operator & \{uh\} Give me one second here. \\
     $U_6$ &Operator & So, could I get your, I need some information. \\
     $U_7$ &Operator & To check this on our computer system, could I get your name please? \\
     $U_8$ & Customer & Yeah, my name is Maria Wilson. \\
     $U_9$ &Operator & Maria \{mhm\}. \\
     $U_{10}$ & Customer & \{uh\} Yeah, that's W I L S O N. \\
     $U_{11}$ &Operator & Thank you for that, and could I get your account number please? \\
     $U_{12}$ & Customer & Yeah, it's \{uh\} two one eight five. \\
     $U_{13}$ & Customer & One nine, six six, four five. \\
     $U_{14}$ &Operator & \{uh\} And your current address. \\
     $U_{15}$ & Customer & Yeah, it's one two three, main street \{uh\} Brick tower, number four five one three. \\
     ... & ... & ... \\ \hline
  \end{tabular}
 \end{center}
\end{table*}

\section{Problem Formulations}
\subsection{Definition of conversational documents}
A conversational document is represented as a sequence of utterance-level information $\bm{U}=\{U^{1},\cdots,U^{T}\}$. The utterance information is composed of speaker information and text information. Thus, we represent the $t$-th utterance as a speaker label $q^{t}$ and text $W^t$. The $t$-th text is represented as a word sequence $W^{t}=\{w_1^{t},\cdots,w_{N^{t}}^{t}\}$. Table 1 shows an example of a conversational document in a contact center dialogue. In this case, the speaker label is either operator or customer.

\subsection{Definition of utterance-level sequential labeling}
In the utterance-level sequential labeling for conversational documents, utterance-level labels $\bm{O}=\{o^{1},\cdots,o^{T}\}$ are estimated from a sequence of utterance-level information $\bm{U}=\{U^{1},\cdots,U^{T}\}$. In this paper, we construct the utterance-level sequential labeling from unlabeled sentences ${\cal D}_{\tt sentences} = \{W^1,\cdots,W^J\}$, unlabeled conversational documents ${\cal D}_{\tt unpair} = \{\bm{U}_1,\cdots,\bm{U}_M\}$, and labeled conversational documents ${\cal D}_{\tt pair} =$ $ \{(\bm{U}_{M+1},\bm{O}_{M+1}),$ $\cdots,$ $(\bm{U}_{M+K},\bm{O}_{M+K})\}$. ${\cal D}_{\tt unpair}$ and ${\cal D}_{\tt sentences}$ are used for the pre-training step, and ${\cal D}_{\tt pair}$ is used for the fine-tuning step. 

\section{Large-Context Conversational Representation Learning}
This section details our proposed large-context conversational representation learning (LC-CRL).

\subsection{Modeling}
Our self-supervised task is to estimate an utterance from all past utterances and all future utterances. In other words, we simultaneously estimate both a speaker label and a word sequence from all conversational contexts. We estimate the predicted probability of the $t$-th utterance from all conversational contexts as 
\begin{multline}
 P(U^{t}|U^{1:t-1},U^{t+1:T}; \bm{\Theta}) = \\
 P(q^{t}|U^{1:t-1},U^{t+1:T}; \bm{\Theta}) \\
 \prod_{n=1}^{N_t} P(w_n^{t}|w_{1:n-1}^{t}, q^{t}, U^{1:t-1},U^{t+1:T}; \bm{\Theta}) ,
\end{multline}
where $\bm{\Theta}$ represents the model parameter of the model. This model is composed of an utterance encoder, a past-context encoder, a future-context encoder, and an utterance decoder. The network structure of the LC-CRL modeling is shown in Fig. 1.

\subsection{Utterance encoder}
In the utterance encoder, a speaker label and all words in an utterance are embedded into a continuous vector using bidirectional long short-term memory recurrent neural networks (LSTM-RNNs) and the self-attention mechanism. The $t$-th utterance's continuous vector is computed by
\begin{equation}
 \bm{S}^{t} = {\tt SelfAttention}(\bm{C}^t; \bm{\theta}_{\tt s}) ,
\end{equation}
where ${\tt SelfAttention}()$ is a function that uses an attention mechanism to summarize several continuous vectors as one continuous vector \cite{lin_iclr2017}; $\bm{\theta}_{\tt s}$ is the trainable parameter. $\bm{C}^t$ is computed from the word continuous vector as
\begin{equation}
 \bm{C}^t = {\tt BLSTM}([{\bm{q}^t}^\top,{\bm{w}^t_1}^\top]^\top,\cdots,[{\bm{q}^t}^\top,{\bm{w}^t_{N^t}}^\top]^\top; \bm{\theta}_{\tt c}) ,
\end{equation}
where ${\tt BLSTM}()$ is a function of bidirectional LSTM-RNNs; $\bm{\theta}_{\tt c}$ is the trainable parameter. The $n$-th word in the $t$-th utterance $\bm{w}^t_{n}$ and the speaker label for the $t$-th utterance $\bm{q}^t$ are defined as
\begin{equation}
 \bm{w}_{n}^t = {\tt EMBED}(w_{n}^t;\bm{\theta}_{\tt w}) ,
\end{equation}
\begin{equation}
 \bm{q}^t = {\tt EMBED}(q^t;\bm{\theta}_{\tt q}) ,
\end{equation}
where ${\tt EMBED}()$ is a linear transformational function that embeds a symbol into a continuous vector; $\bm{\theta}_{\tt w}$ and $\bm{\theta}_{\tt q}$ are the trainable parameters. Note that we can utilize pre-trained word representations including both continuous bag-of-words and ELMo.

\subsection{Past-context encoder}
The past-context encoder embeds continuous vectors of the past utterances into a continuous vector using an utterance-level forward LSTM-RNN. The continuous vector that embeds utterances from the $1$-st utterance to the $t-1$-th utterance is defined as
\begin{equation}
 \begin{split}
 \bm{L}^{t} & = \overrightarrow{\tt LSTM}(\bm{S}^{1},\cdots,\bm{S}^{t-1}; \bm{\theta}_{\tt l})\\
 & = \overrightarrow{\tt LSTM}(\bm{S}^{t-1},\bm{L}^{t-1}; \bm{\theta}_{\tt l}) ,
 \end{split}
\end{equation}
where $\overrightarrow{\tt LSTM}()$ is a function of the forward LSTM-RNN; $\bm{\theta}_{\tt l}$ is the trainable parameter.

\begin{figure}[t]
\begin{center}
\includegraphics[width=85mm]{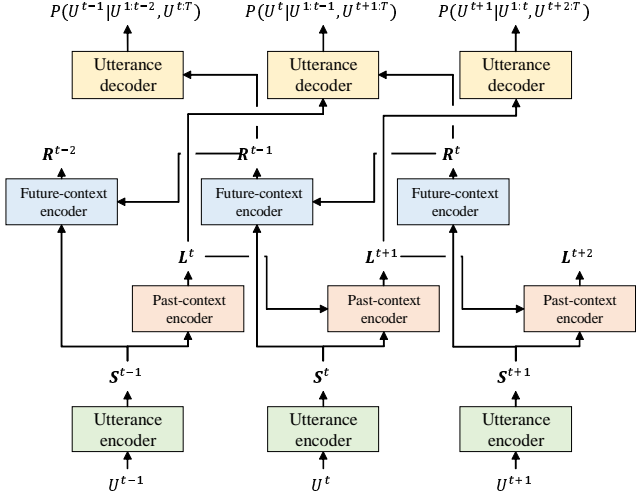}
\end{center}
\vspace{-4mm} 
\caption{Network structure for LC-CRL modeling.}
\end{figure}

\subsection{Future-context encoder}
The future-context encoder embeds the continuous vectors of the future utterances into a continuous vector using an utterance-level backward LSTM-RNN. The continuous vector that embeds utterances from the $t+1$-th utterance to the $T$-th utterance is defined as
\begin{equation}
 \begin{split}
 \bm{R}^{t} & = \overleftarrow{\tt LSTM}(\bm{S}^{t+1},\cdots,\bm{S}^{T}; \bm{\theta}_{\tt r})\\
 & = \overleftarrow{\tt LSTM}(\bm{S}^{t+1},\bm{R}^{t+1}; \bm{\theta}_{\tt r}) , 
 \end{split}
\end{equation}
where $\overleftarrow{\tt LSTM}()$ is a function of the backward LSTM-RNN; $\bm{\theta}_{\tt r}$ is the trainable parameter.

\subsection{Utterance decoder}
The utterance decoder estimates both a speaker label and the words in an utterance by leveraging continuous vectors output by both the past-context encoder and the future-context encoder. A predicted probability of the $t$-th speaker label is formulated as 
\begin{multline}
 P(q^{t}|U^{1:t-1},U^{t+1:T}; \bm{\Theta}) = \\ {\tt SOFTMAX}([{\bm{L}^{t}}^\top,{\bm{R}^{t}}^\top]^\top; \bm{\theta}_{\tt y}) , 
\end{multline}
where ${\tt SOFTMAX}()$ is a linear transformational function with softmax activation; $\bm{\theta}_{\tt y}$ is the trainable parameter. In addition, the generative probabilities of words are estimated on the basis of auto-regressive generative modeling. The generative probabilities of the $n$-th word in the $t$-th utterance are produced by
\begin{multline}
 P(w_n^{t}|w_1^{t},\cdots,w_{n-1}^{t}, q^{t}, U^{1:t-1},U^{t+1:T}; \bm{\Theta}) \\
 = {\tt SOFTMAX}(\bm{v}_{n}^{t}; \bm{\theta}_{\tt d}) ,
\end{multline}
where $\bm{\theta}_{\tt d}$ is the trainable parameter, and $\bm{v}_n^t$, which represents context information, is given by
\begin{equation}
 \bm{v}_{n}^{t} = \overrightarrow{\tt LSTM}( [{\bm{w}_{n-1}^{t}}^\top,{\bm{q}^{t}}^\top,{\bm{L}^{t}}^\top,{\bm{R}^{t}}^\top]^\top ,\bm{v}_{n-1}^{t}; \bm{\theta}_{\tt v}) ,
\end{equation}
where $\bm{\theta}_{\tt v}$ is the trainable parameter. Thus, the context information includes not only long-range linguistic contexts within an utterance but also long-range past and future contexts beyond the utterance boundaries.

\subsection{Optimization}
The proposed method utilizes unlabeled conversational documents ${\cal D}_{\tt unpair} = \{\bm{U}_1,\cdots,\bm{U}_M\}$ for learning networks to extract conversational document representations from the conversational document. Here, model parameters are represented as $\bm{\Theta}=\{\bm{\theta}_{\tt s},\bm{\theta}_{\tt c},\bm{\theta}_{\tt w},\bm{\theta}_{\tt q}$, $\bm{\theta}_{\tt l}, \bm{\theta}_{\tt r},\bm{\theta}_{\tt y},\bm{\theta}_{\tt d},\bm{\theta}_{\tt v}\}$. The model parameters can be optimized by
\begin{multline}
 \hat{\bm{\Theta}} = \argmin_{\bm{\Theta}} - \sum_{\bm{U} \in {\cal D}_{\tt unpair}} \sum_{t=1}^{T} \\
 \log P(U^{t}|U^{1:t-1},U^{t+1:T}; \bm{\Theta}) .
\end{multline}
The optimization is achieved by means of a mini-batch stochastic gradient decent algorithm with conversation-level mini-batches.

We can also apply a two-stage representation learning method in which context-dependent word representations are trained in the first stage and context-dependent utterance representations, LC-CRL, are trained in the second stage. For the first stage, we construct ELMo \cite{peters_naacl2018} from ${\cal D}_{\tt sentences} = \{W^1,\cdots,W^J\}$. This two-stage representation learning method enables us to utilize not only unlabeled conversational documents but also unlabeled sentences.

\section{Utterance-Level Sequential Labeling with LC-CRL}
This section describes a method to build an utterance-level sequential labeling model based on a speaker-aware hierarchical bidirectional long short-term memory recurrent neural network conditional random field (SA-H-BLSTM-CRF) with LC-CRL.

\subsection{Modeling}
In SA-H-BLSTM-CRF-based utterance-level sequential labeling, we model the conditional probabilities of utterance-by-utterance labels $\bm{O}$ given an entire conversational document $\bm{U}$. The utterance-by-utterance labels are estimated by
\begin{equation}
 \hat{\bm{O}} = \argmax_{\bm{O}} P(\bm{O}|\bm{U}; \bm{\Lambda}) ,
\end{equation}
where $\bm{\Lambda}$ indicates the model parameters of SA-H-BLSTM-CRF. This model is composed of an utterance encoder, a past-context encoder, a future-context encoder, and a CRF-based classifier. Fig. 2 shows the network structure of the SA-H-BLSTM-CRF. In comparison to the network in Fig. 1, the structure of SA-H-BLSTM-CRF is pretty similar to LC-CRL modeling. There are two key differences between these two network structures: one, the type of network output, and two, how the outputs from the past-context encoder and the future-context encoder are aggregated. Note that the utterance encoder, past-context encoder, and future-context encoder have the same structures between SA-H-BLSTM-CRF and LC-CRL modeling.

\begin{figure}[t]
\begin{center}
\includegraphics[width=85mm]{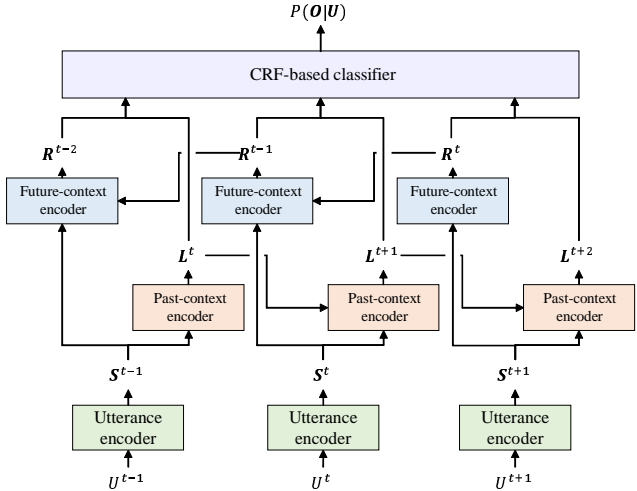}
\end{center}
\vspace{-4mm} 
\caption{Network structure for SA-H-BLSTM-CRF.}
\end{figure}

\subsection{CRF-based classifier}
The CRF-based classifier estimates the predicted probabilities of utterance-by-utterance labels $\bm{O}$ from continuous representations generated from both the past-context encoder and the future-context encoder. The predicted probabilities are computed from 
\begin{equation}
P(\bm{O}|\bm{U};\bm{\Lambda}) = \frac{\prod_{t=1}^T \exp \phi(o^{t-1},o^{t},\bm{y}^{t};\bm{\theta}_{\tt o})}{\sum_{\bar{\bm{O}}} \prod_{t=1}^T \exp \phi(\bar{o}^{t-1},\bar{o}^{t},\bm{y}^{t};\bm{\theta}_{\tt o})} \displaystyle ,
\end{equation}
where $\phi()$ is the linear transformational function in the linear CRF layer and $\bm{\theta}_{\tt o}$ is the model parameter. $\bm{y}^t$ refers to the $t$-th continuous representations generated from both the past-context encoder and the future-context encoder. $\bm{y}^t$ is constituted by 
\begin{equation}
 \bm{y}^{t} = [{\bm{L}^{t+1}}^\top,{\bm{R}^{t-1}}^\top]^\top .
\end{equation}
Note that the CRF-based classifier uses the outputs of both the past-context encoder and future-context encoder differently from the utterance decoder in the LC-CRL.

\subsection{Optimization}
To optimize the utterance-level sequential labeling based on SA-H-BLSTM-CRF, the model parameters $\bm{\Lambda}=\{\bm{\theta}_{\tt s},\bm{\theta}_{\tt c},\bm{\theta}_{\tt w}$, $\bm{\theta}_{\tt q},\bm{\theta}_{\tt l},\bm{\theta}_{\tt r},\bm{\theta}_{\tt o}\}$ are optimized from labeled conversational documents ${\cal D}_{\tt pair}$. Note that $\{\bm{\theta}_{\tt s},\bm{\theta}_{\tt c},\bm{\theta}_{\tt w},\bm{\theta}_{\tt q},\bm{\theta}_{\tt l},\bm{\theta}_{\tt r}\}$ are the pre-trained parameters when LC-CRL-based pre-training is performed. The model parameters can be optimized by 
\begin{equation}
 \hat{\bm{\Lambda}} = \argmin_{\bm{\Lambda}}
 - \sum_{(\bm{U},\bm{O}) \in {\cal D}_{\tt pair}} \log P(\bm{O}|\bm{U};\bm{\Lambda}) .
\end{equation}
The optimization is performed using the mini-batch stochastic gradient decent algorithm with conversation-level mini-batches.

\setcounter{table}{0}
\begin{table}[t!]
 \caption{Experimental datasets.}
 \begin{center}
  \begin{tabular}{|l|rrr|} \hline 
    Business & No. of & No. of & No. of \\ 
    type & calls & utterances & words \\ \hline \hline
    Finance & 59 & 6,081 & 55,933 \\ 
    Internet provider & 57 & 3,815 & 47,668 \\ 
    Government unit & 73 & 5,617 & 48,998 \\ 
    Mail-order & 56 & 4,938 & 46,574 \\ 
    PC repair & 55 & 6,263 & 55,101 \\ 
    Mobile phone & 61 & 5,738 & 51,061 \\  \hline \hline
    All & 361 & 32,452 & 305,351 \\ \hline
  \end{tabular}
 \end{center}
\end{table}

\begin{table*}[t!]
  \caption{Experimental results in terms of classification accuracy (\%) and F-measure (\%) with respect to output labels.}
 \label{}
 \begin{center}
  \begin{tabular}{|lll|r|rrrrr|} \hline
    & Model & Self-supervised learning & Accuracy (\%) & \multicolumn{5}{|c|}{F-measure (\%)} \\
    &  & & & C1 & C2 & C3 & C4 & C5 \\ \hline \hline
    1. & H-BLSTM-CRF & -- & 83.8 & 77.4 & 55.2 & 88.3 & 83.6 & 80.1 \\ 
    2. & SA-H-BLSTM-CRF & -- & 85.0 & 78.7 & 58.5 & 90.6 & 85.6 & 81.3 \\
    3. & SA-H-BLSTM-CRF & ELMo & 88.2 & 83.5 & 63.3 & 91.1 & 88.3 & 82.9 \\ 
    4. & SA-H-BLSTM-CRF & STV & 87.2 & 80.2 & 60.6 & 91.6 & 88.2 & 83.8 \\
    5. & SA-H-BLSTM-CRF & ELMo+STV & 88.7 & 85.2 & 68.4 & 92.7 & 90.6 & 85.8 \\
    6. & SA-H-BLSTM-CRF & LC-CRL & 89.6 & 84.0 & 64.3 & 91.6 & 89.6 & 84.2 \\
    7. & SA-H-BLSTM-CRF & ELMo+LC-CRL & {\bf 90.4} & {\bf 85.7} & {\bf 70.4} & {\bf 93.4} & {\bf 92.1} & {\bf 87.3} \\ \hline
  \end{tabular}
 \end{center}
 \vspace{-5mm}
\end{table*}

\section{Experiments}
\subsection{Datasets}
We performed our experiments using simulated contact center dialogue datasets consisting of 361 labeled conversational documents in six business fields. One dialogue means one telephone call between one operator and one customer. All utterances were manually transcribed. Each dialogue was automatically divided into speech units using speech activity detection. We manually annotated five call scenes: C1: opening, C2: requirement confirmation, C3: response, C4: customer confirmation, and C5: closing \cite{masumura_apsipa2018}. Detailed setups are shown in Table 2, where No. of calls, No. of utterances, and No. of words indicate the number of calls, words, and utterances with respect to the call scenes, respectively. We also prepared 4,000 unlabeled conversational documents collected from various contact centers for self-supervised learning and prepared an additional 500 million sentences collected from the Web. 

\subsection{Setups}
The evaluation involved 6-fold cross validation open to business type, where five business types were used for training and the remaining one for testing. As baselines, we constructed H-BLSTM-CRF and speaker-aware H-BLSTM-CRF (SA-H-BLSTM-CRF) models. The network configurations in these models were unified as follows. We defined the word vector representation as a 512-dimensional vector and the speaker vector representation as a 32-dimensional vector. Words that appeared only once in the training datasets were treated as unknown words. We used two-layer BLSTM-RNNs with 512 units for the utterance encoder and two-layer LSTM-RNNs with 512 units for both the past-context encoder and future-context encoder. Dropout with the rate set to 0.2 was used for ${\tt BLSTM()}$ and ${\tt LSTM()}$. In addition, we constructed ELMo \cite{peters_naacl2018}, skip-thought vector (STV) \cite{kiros_nips2015}, and LC-CRL models for comparison. ELMo was trained only from sentences collected from the Web, and STV and LC-CRL were trained from unpaired conversational documents. Note that we examined two-stage training individually for STV and LC-CRL using the pre-trained ELMo.

For training, the mini-batch size was set to five conversational documents. The optimizer was Adam with the default settings. Note that a part of the training sets was used as the datasets utilized for early stopping. We constructed five models by varying the initial parameters and performed our evaluations using the model that had the lowest validation loss for individual setups.

\subsection{Results}
Table 3 shows the experimental results in terms of classification accuracy (\%) and F-measure (\%) with respect to output call scene labels. We can see that line 2, which captures not only textual information but also speaker information, outperformed line 1, which only uses textual information. This indicates that considering who spoke what in what order is important for utterance-level sequential labeling for conversational documents. Next, lines 3--7 yielded a higher performance than line 2. This shows that self-supervised learning can improve the utterance-level sequential labeling performance. Our proposed method, LC-CRL, outperformed both ELMo and STV, because LC-CRL can learn the relationships between utterances. The highest performance was achieved by line 7, which utilizes both ELMo and LC-CRL for two-stage self-supervised learning. This confirms that the proposed method that utilizes both unlabeled sentences and unlabeled conversational documents for self-supervised learning is an effective approach.

Fig. 3 shows the classification accuracy results when the data size of paired datasets was varied. When the paired data size was small, the classification performance of SA-H-BLSTM-CRF+ELMo was not high. This is because ELMo-based self-supervised learning cannot capture relationships between utterances. In contrast, SA-H-BLSTM-CRF+ELMo+LC-CRL could attain a higher performance even when the labeled datasets were small. These results demonstrate that our proposed LC-CRL is effective for the utterance-level sequential labeling of conversational documents.

\begin{figure}[t]
  \begin{center}
    \includegraphics[width=85mm]{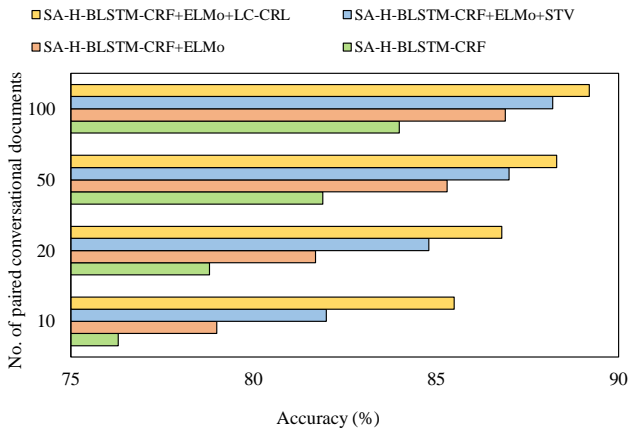}
  \end{center}
  \vspace{-6mm}
  \caption{Results when using different size of paired datasets.}
\end{figure}

\section{Conclusions}
In this paper, we proposed large-context conversational representation learning (LC-CRL), a self-supervised learning method specialized for conversational documents. The key strength of LC-CRL is that it can leverage unlabeled conversational documents for acquiring networks to generate context-dependent utterance representations. The pre-trained networks can then be used for the utterance-level sequential labeling networks. Experiments on scene segmentation tasks using contact center datasets showed that the proposed method yielded a better performance than methods without self-supervised learning and those with conventional self-supervised learning. We also demonstrated that the proposed method could show superior performance even when few paired datasets were used.

\end{document}